\documentclass[11pt,a4paper]{article}
\usepackage[]{emnlp2021}
\usepackage{times}
\usepackage{latexsym}

\usepackage{microtype}

\usepackage{times}
\usepackage{latexsym}

\usepackage{algorithm,algpseudocode,amsfonts,amsmath,balance,booktabs,color,epsfig,graphics,graphicx,multirow,paralist,pifont,setspace,subfigure,url,wrapfig,xspace}

\definecolor{darkred}{rgb}{0.8,0.0,0.0}
\definecolor{darkblue}{rgb}{0.0,0.0,0.7}
\definecolor{darkgreen}{rgb}{0.0,0.5,0.0}
\definecolor{darkorange}{rgb}{1.0,0.4,0.0}
\newcommand{\textred}[1]{\textcolor{darkred}{#1}}
\newcommand{\textblue}[1]{\textcolor{darkblue}{#1}}
\newcommand{\textgreen}[1]{\textcolor{darkgreen}{#1}}
\newcommand{\textorange}[1]{\textcolor{darkorange}{#1}}

\title{\vspace*{-0.5in}
{\small \hfill EMNLP-Eval4NLP Workshop 2021} \\
\vspace{0.35in}
Validating Label Consistency in NER Data Annotation}

\author{{\bf Qingkai Zeng$^{\dag}$, Mengxia Yu$^{\dag}$, Wenhao Yu$^{\dag}$, Tianwen Jiang$^{\ddag}$,} \\
{\bf Meng Jiang$^{\dag}$} \\
$\dag$University of Notre Dame, Notre Dame, IN, USA\\ 
$\ddag$Harbin Institute of Technology, Harbin, Heilongjiang, China\\
{\tt $\dag$\{qzeng, myu2, wyu1, mjiang2\}@nd.edu} \\
{\tt $\ddag$\{twjiang\}@ir.hit.edu.cn}
}
\date{}

\begin{document}
\maketitle
\begin{abstract}
Data annotation plays a crucial role in ensuring your named entity recognition (NER) projects are trained with the correct information to learn from. Producing the most accurate labels is a challenge due to the complexity involved with annotation. Label inconsistency between multiple subsets of data annotation (e.g., training set and test set, or multiple training subsets) is an indicator of label mistakes. In this work, we present an empirical method to explore the relationship between label (in-)consistency and NER model performance. It can be used to validate the label consistency (or catch the inconsistency) in multiple sets of NER data annotation. In experiments, our method identified the label inconsistency of test data in SCIERC and CoNLL03 datasets (with 26.7\% and 5.4\% label mistakes). It validated the consistency in the corrected version of both datasets.
\end{abstract}

\section{Introduction}
\label{sec:introduction}
\begin{table*}[t]
\centering
\caption{Three examples to compare original and corrected annotation in the test set of the SCIERC dataset. If the annotation on the test set consistently followed the ``codebook'' that was used to annotate training data, the entities in the first two examples would be labelled as ``Task'' (not ``Method'') for sure.}
\label{tab:stoa-result}
\scalebox{0.82}{%
\linespread{1.08}
\begin{tabular}{p{9.2cm}|p{9.3cm}}
\toprule
\centering{\textbf{Original Examples}} & \multicolumn{1}{c}{\textbf{Corrected Examples}} \\ \hline
\normalsize{Starting from a DP-based solution to the \textbf{\textred{[traveling salesman problem]}}}\small{\textbf{\textgreen{Method}}}\normalsize{, we present a novel technique ...} & \normalsize{Starting from a DP-based solution to the \textred{\textbf{[traveling salesman problem]}}}\small{\textbf{\textblue{Task}}}\normalsize{, we present a novel technique ...} \\ \hline
\normalsize{FERRET utilizes a novel approach to \textbf{\textred{[Q/A]}}}\small{\textbf{\textgreen{Method}}} \normalsize{known as predictive questioning which attempts to identify ...} & \normalsize{FERRET utilizes a novel approach to \textbf{\textred{[Q/A]}}}\small{\textbf{\textblue{Task}}} \normalsize{known as predictive questioning which attempts to identify ...} \\ \hline
\normalsize{The goal of this work is the enrichment of \textbf{\textred{[human-machine interactions]}}}\small{\textbf{\textblue{Task}}} \normalsize{in a natural language environment.} & \normalsize{The goal of this work is the \textbf{\textred{[enrichment of human-machine interactions]}}}\small{\textbf{\textblue{Task}}} \normalsize{in a natural language environment.} \\
\bottomrule
\end{tabular}}
\label{tab:case-study}
\end{table*}

\begin{figure*}[t]
\centering
{\includegraphics[width=\textwidth]{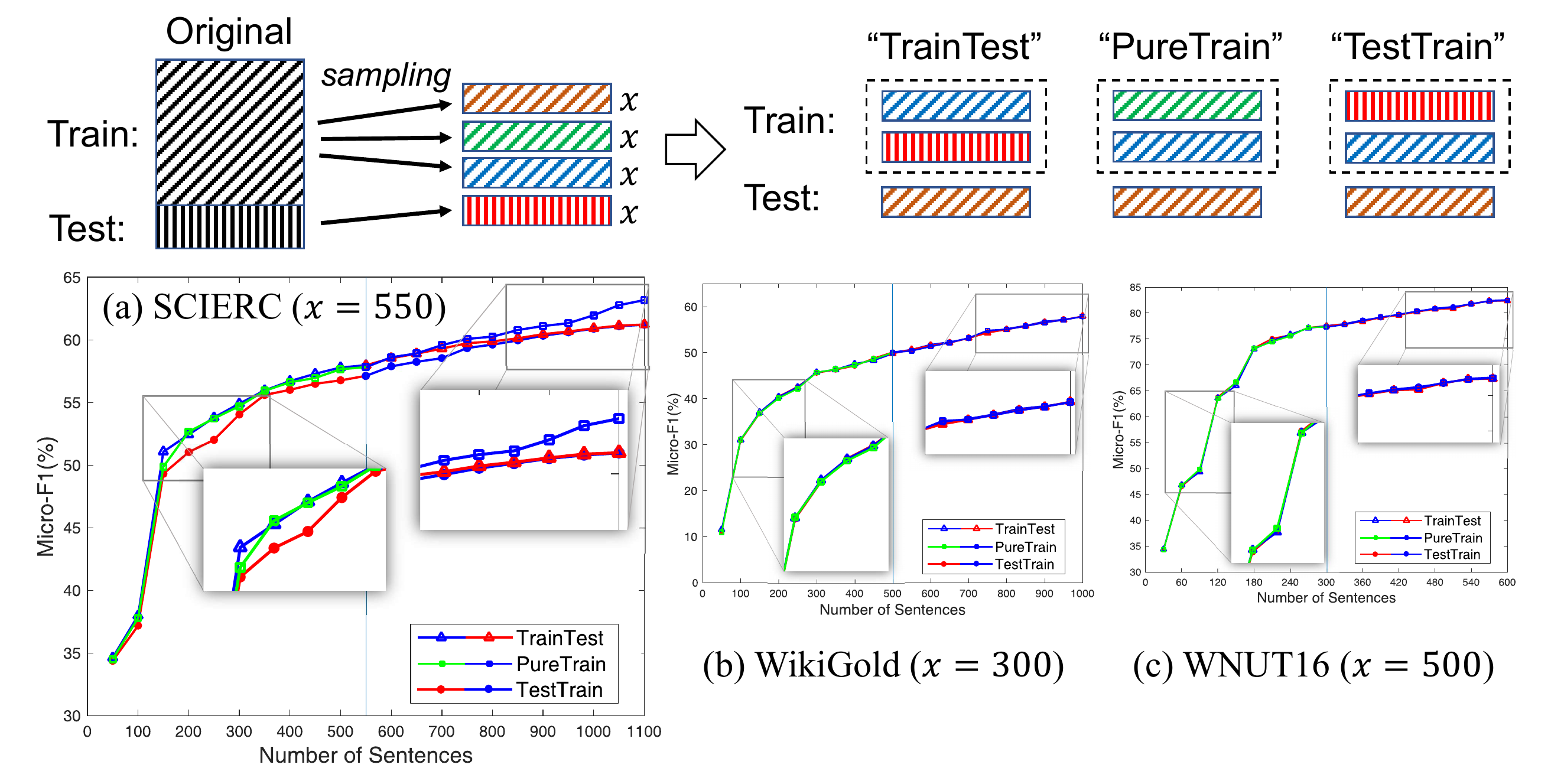}}
\vspace{-0.3in}
\caption{\emph{Identifying label inconsistency of test set with training set:} We sample \emph{three} exclusive subsets (of size $x$) from the training set (\textorange{orange}, \textgreen{green}, and \textblue{blue}). We use one subset as the \emph{new} test set (\textorange{orange}). We apply the \textsc{SCIIE} NER model on the new test set. We build three \emph{new} training sets: \emph{i)} ``TrainTest'' (\textblue{blue}-\textred{red}), \emph{ii)} ``PureTrain'' (\textgreen{green}-\textblue{blue}), \emph{iii)} ``TestTrain'' (\textred{red}-\textblue{blue}). Results on SCIERC show that the test set (\textred{red}) is \textit{less predictive} of training samples (\textorange{orange}) than the training set itself (\textblue{blue} or \textgreen{green}). This was not observed on two other datasets.}
\label{fig:id_test_mistake}
\end{figure*}

Named entity recognition (NER) is one of the foundations of many downstream tasks such as relation extraction, event detection, and knowledge graph construction. NER models require vast amounts of labeled data to learn and identify patterns that humans cannot continuously. It is really about getting accurate data to train the models. When end-to-end neural models achieve excellent performance on NER in various domains \cite{lample2016neural,liu2018empower,luan2018multi,zeng2020tri,zeng2021enhancing}, building useful and challenging NER benchmarks, such as CoNLL03, WNUT16, and SCIERC, contributes significantly to the research community.

Data annotation plays a crucial role in building benchmarks and ensuring NLP models are trained with the correct information to learn from~\cite{luan2018multi,jiang2020biomedical,yu2020identifying}. Producing the necessary annotation from any asset at scale is a challenge, mainly because of the complexity involved with annotation. Getting the most accurate labels demands time and expertise.

Label mistakes can hardly be avoided, especially when the labeling process splits the data into multiple sets for distributed annotation. The mistakes cause label inconsistency between subsets of annotated data (e.g., training set and test set or multiple training subsets). For example, in the CoNLL03 dataset \cite{sang2003introduction}, a standard NER benchmark that has been cited over 2,300 times, label mistakes were found in 5.38\% of the test set \cite{wang2019crossweigh}. Note that the state-of-the-art results on CoNLL03 have achieved an F1 score of $\sim.93$. So even if the label mistakes make up a tiny part, they cannot be negligible when researchers are trying to improve the results further. In the work of Wang \emph{et al.}, five annotators were recruited to correct the label mistakes. Compared to the original test set results, the corrected test set results are more accurate and stable.

However, two critical issues were not resolved in this process: \emph{i)} How to identify label inconsistency between the subsets of annotated data? \emph{ii)} How to validate that the label consistency was recovered by the correction?

Another example is SCIERC \cite{luan2018multi} (cited $\sim$50 times) which is a multi-task (including NER) benchmark in AI domain. It has 1,861 sentences for training, 455 for dev, and 551 for test. When we looked at the false predictions given by \textsc{SCIIE} which was a multi-task model released along with the SCIERC dataset, we found that as many as 147 (26.7\% of the test set) sentences were not properly annotated. (We also recruited five annotators and counted a mistake when all the annotators report it.) Three examples are given in Table~\ref{tab:case-study}: two of them have wrong entity types; the third has a wrong span boundary. As shown in the experiments section, after the correction, the NER performance becomes more accurate and stable.

\begin{figure*}[t]
\centering
\includegraphics[width=1.0\linewidth]{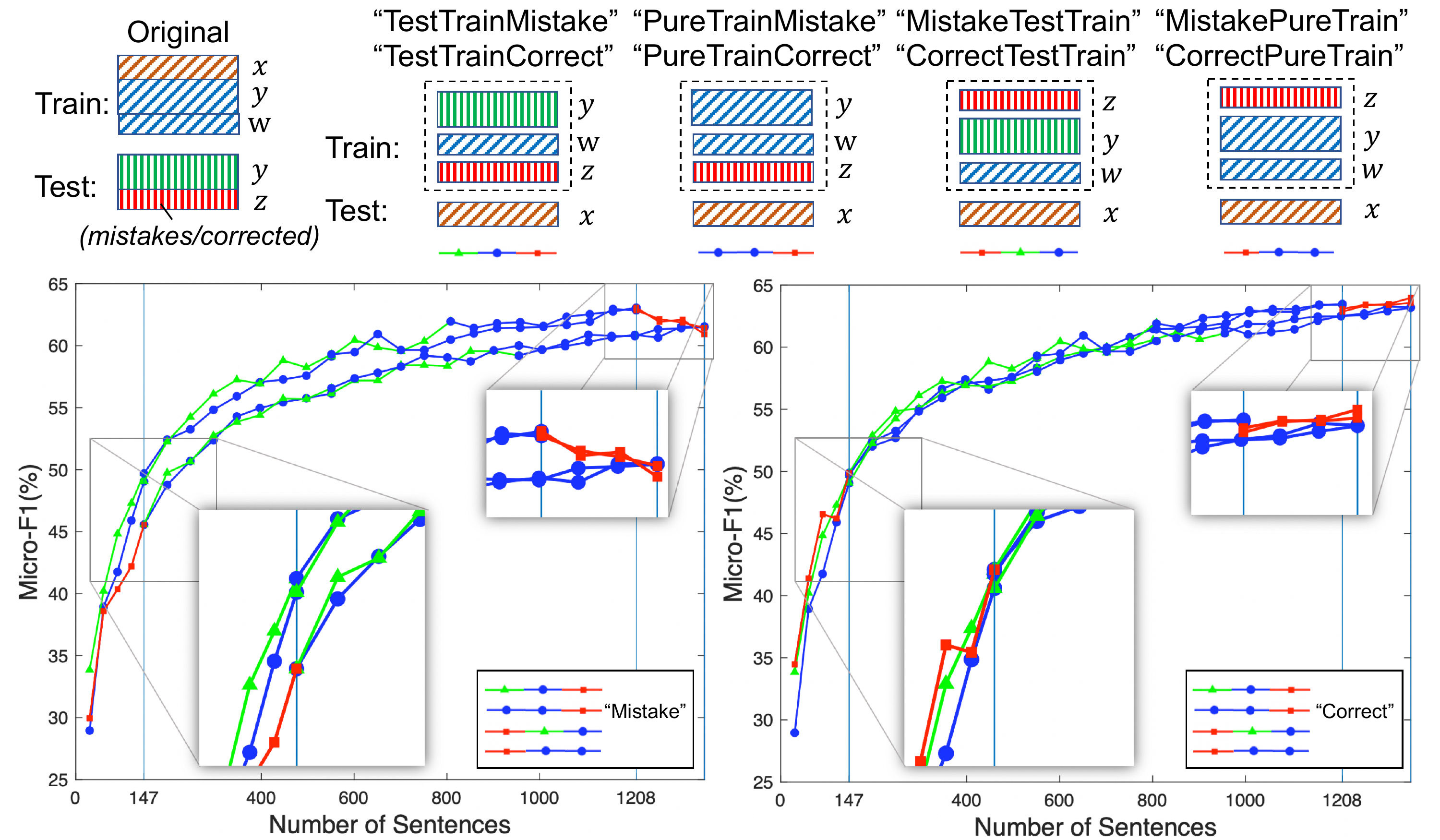}
\vspace{-0.2in}
\caption{\textit{Validating label consistency in corrected test set}:
We corrected $z$ of $y+z$ sentences in the test set. We sampled \emph{three} exclusive subsets of size $x$, $y$, and $w$ from the training set. We use the first subset (of size $x$) as the \emph{new} test set. We build four \emph{new} training sets as shown in the figure and feed them into the \textsc{SCIIE} model (at the top of the figure). Results show that the label mistakes (\textred{red} parts of the curves on the left) do hurt the performance no matter fed at the beginning or later; and the corrected test set performs as well as the training set (on the right).}
\label{fig:val_test_correct}
\end{figure*}

Besides the significant correction on the SCIERC dataset, our contributions in this work are as follows: \emph{i)} an empirical, visual method to identify the label inconsistency between subsets of annotated data (see Figure~\ref{fig:id_test_mistake}), \emph{ii)} a method to validate the label consistency of corrected data annotation (see Figure~\ref{fig:val_test_correct}). Experiments show that they are effective on the CoNLL03 and SCIERC datasets.

\section{Proposed Methods}
\label{sec:method}

\subsection{A method to identify label inconsistency}

Suppose the labeling processes on two parts of annotated data were consistent. They are likely to be equivalently predictive of each other. In other words, if we train a model with a set of samples from either part $A$ or part $B$ to predict a different set from part $A$, the performance should be similar.

Take SCIERC as an example. We were wondering whether the labels in the test set were consistent with those in the training set. Our method to identify the inconsistency is presented in Figure~\ref{fig:id_test_mistake}.

We sample \emph{three} exclusive subsets (of size $x$) from the training set. We set $x = 550$ according to the size of the original test set. We use one of the subsets as the \emph{new} test set. Then we train the \textsc{SCIIE} NER model \cite{luan2018multi} to perform on the new test set. We build three \emph{new} training sets to feed into the model:
\begin{compactitem}
    \item ``TrainTest'': first fed with one training subset and then the original test set;
    \item ``PureTrain'': fed with two training subsets;
    \item ``TestTrain'': first fed with the original test set and then one of the training subsets.
\end{compactitem}

Results show that ``TestTrain'' performed the worst at the early stage because the quality of the original test set is not reliable. In ``TrainTest'' the performance no longer improved when the model started being fed with the original test set. ``PureTrain'' performed the best. All the observations conclude that the original test set is less predictive of training samples than the training set itself. It may be due to the issue of label inconsistency. Moreover, we do not have such observations on two other datasets, WikiGold and WNUT16.

\subsection{A method to validate label consistency after correction}

After we corrected the label mistakes, how could we empirically validate the recovery of label consistency? Again, we use a subset of training data as the new test set. We evaluate the predictability of the original wrong test subset, the corrected test subset, and the rest of the training set. We expect to see that the wrong test subset delivers weaker performance and the other two sets make comparable good predictions. Figure~\ref{fig:val_test_correct} illustrates this idea.

Take SCIERC as an example. Suppose we corrected $z$ of $y+z$ sentences in the test set. The original wrong test subset (``Mistake'') and the corrected test subset (``Correct'') are both of size $z$. Here $z = 147$ and the original good test subset $y = 404$ (``Test''). We sampled \emph{three} exclusive subsets of size $x$, $y$, and $w = 804$ from the training set (``Train''). We use the first subset (of size $x$) as the \emph{new} test set. We build four \emph{new} training sets and feed into the \textsc{SCIIE} model. Each new training set has $y+w+z = 1,355$ sentences.
\begin{compactitem}
    \item ``TestTrainMistake''/``TestTrainCorrect'': the original good test subset, the third sampled training subset, and the original wrong test subset (or the corrected test subset);
    \item ``PureTrainMistake''/``PureTrainCorrect'': the second and third sampled training subsets and the original wrong test subset (or the corrected test subset);
    \item ``MistakeTestTrain''/``CorrectTestTrain'': the original wrong test subset (or the corrected test subset), the original good test subset, and the third sampled training subset;
    \item ``MistakePureTrain''/``CorrectPureTrain'': the original wrong test subset (or the corrected test subset) and the second and third sampled training subsets.
\end{compactitem}

Results show that the label mistakes (i.e., original wrong test subset) hurt the model performance whenever being fed at the beginning or later. The corrected test subset delivers comparable performance with the original good test subset and the training set. This demonstrates the label consistency of the corrected test set with the training set.

\section{Experiments}
\label{sec:experiment}
\begin{figure}[t]
\centering
\subfigure[Original with label mistakes]
{\includegraphics[width=0.45\textwidth]{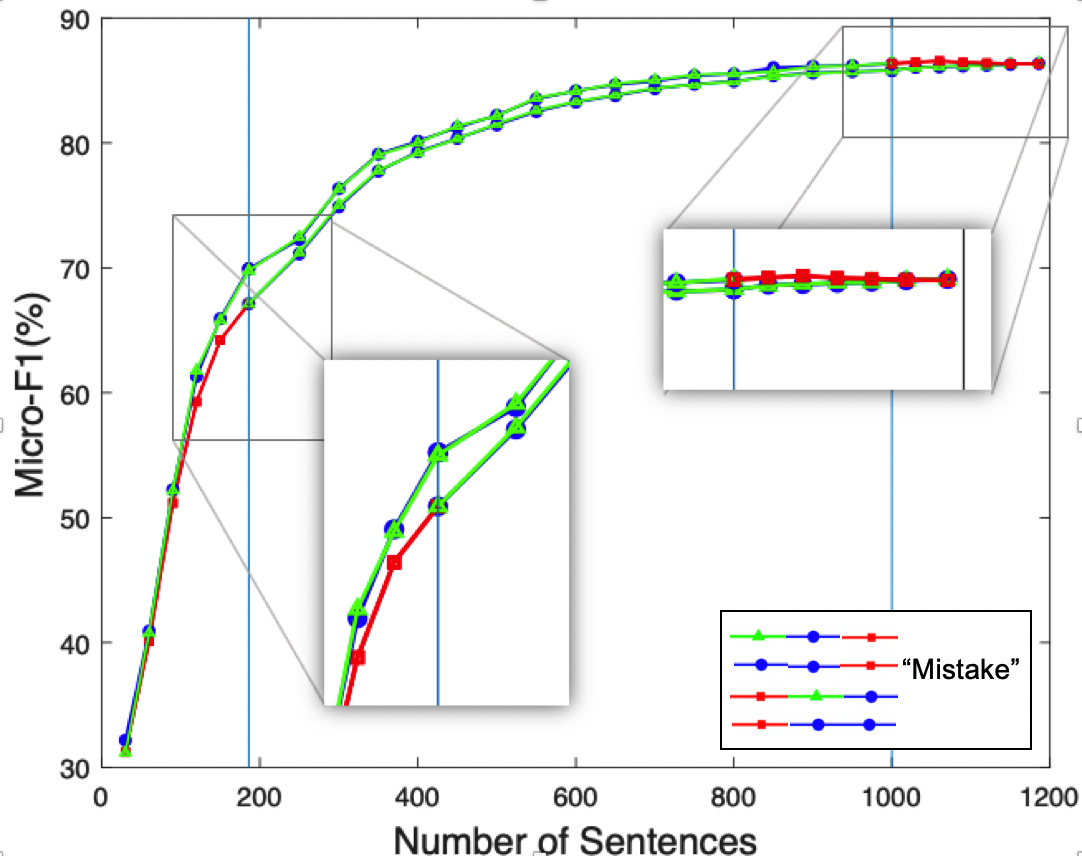}\label{fig:conll03_original}}
\subfigure[Corrected]
{\includegraphics[width=0.45\textwidth]{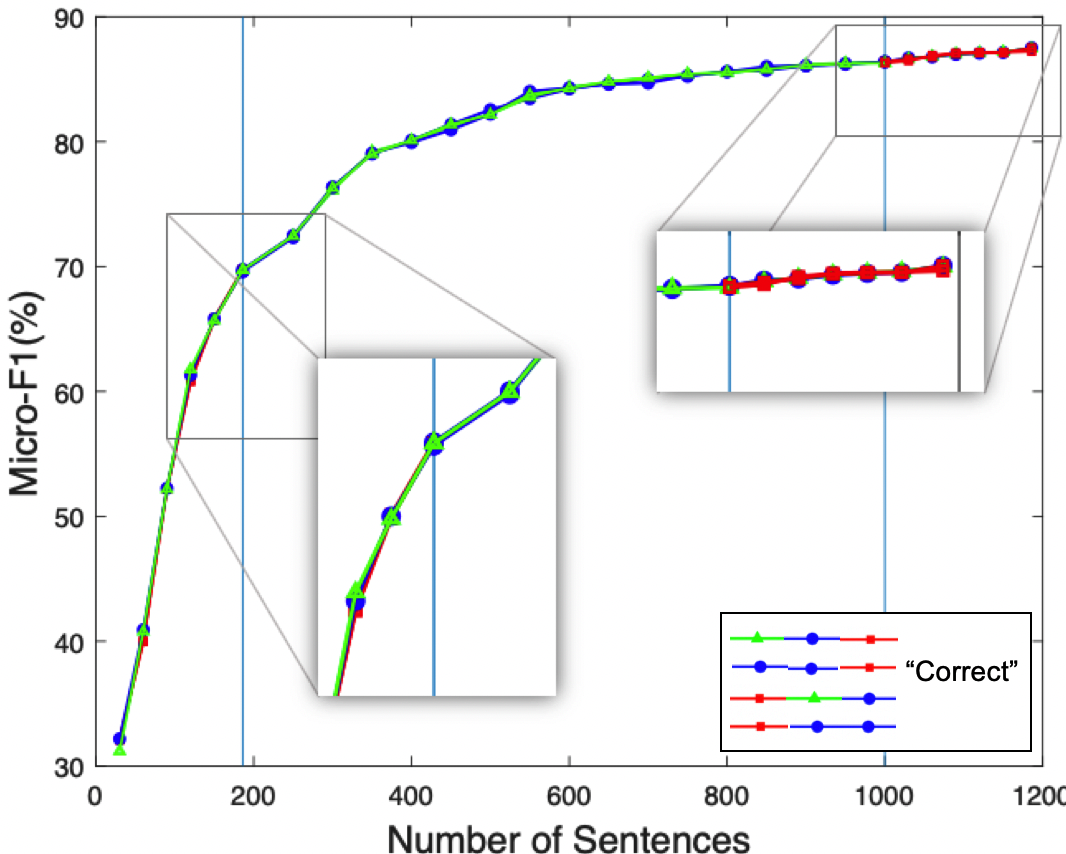}\label{fig:conll03_revised}}
\vspace{-0.1in}
\caption{Identifying label inconsistency and validating the consistency in the original \& corrected CoNLL03.}
\label{fig:val_conll03}
\vspace{-0.2in}
\end{figure}

\subsection{Results on SCIERC}

The visual results of the proposed methods have been presented in Section~\ref{sec:method}. Here we deploy five state-of-the-art NER models to investigate their performance on the corrected SCIERC dataset. The NER models are BiLSTM-CRF \cite{lample2016neural}, LM-BiLSTM-CRF \cite{liu2018empower}, single-task and multi-task SCIIE \cite{luan2018multi}, and multi-task DyGIE \cite{luan2019general}.

\begin{table}[t]
\caption{Five NER models perform consistently better on the corrected SCIERC than on the original dataset.}
\label{tab:Baseline}
\scalebox{0.8}{%
\setlength{\tabcolsep}{0.9mm}{
\begin{tabular}{l|ccc|ccc}
\toprule
\multicolumn{1}{l|}{\multirow{2}{*}{\textbf{Method}}} & \multicolumn{3}{c|}{\textbf{Corrected SCIERC}} & \multicolumn{3}{c}{\textbf{Original SCIERC}} \\
\multicolumn{1}{c|}{} & P & R & F1 & P & R & F1 \\ \hline
{BiLSTM-CRF}  & 58.35 & 47.95 & \textbf{52.64} & 56.13 & 48.07 & \underline{51.79} \\
{LM-BiLSTM-CRF} & 62.78 & 58.20 & \textbf{60.40} & 59.15 & 57.15 & \underline{58.13} \\
{SCIIE-single}  & 71.20 & 62.88 & \textbf{66.79} & 65.77 & 60.90 & \underline{63.24} \\ \hline
{SCIIE-multi} & 72.66 & 63.22 & \textbf{67.61} & 67.66 & 61.72 & \underline{64.56} \\
{DyGIE-multi} & 69.64 & 67.02 & \textbf{68.31} & 65.09 & 65.28 & \underline{65.18} \\ 
 \bottomrule
\end{tabular}}}
\end{table}

As shown in Table \ref{tab:Baseline}, all NER models deliver better performance on the corrected SCIERC than the original dataset. So the training set is more consistent with the fixed test set than the original wrong test set. In future work, we will explore more baselines in the leaderboard.

\subsection{Results on CoNLL03}

Based on the correction contributed by \cite{wang2019crossweigh}, we use the proposed method to justify label inconsistency though the label mistakes take ``only'' 5.38\%. It also validates the label consistency after recovery. Figure~\ref{fig:conll03_original} shows that starting with the wrong labels in the original test set makes the performance worse than starting with the training set or the good test subset. After label correction, this issue is fixed in Figure~\ref{fig:conll03_revised}.


\section{Related Work}
\label{sec:related_work}
NER is typically cast as a sequence labeling problem and solved by models integrate LSTMs, CRF, and language models \cite{lample2016neural,liu2018empower,zeng2019faceted,zeng2020tri}. Another idea is to generate span candidates and predict their type. Span-based models have been proposed with multi-task learning strategies~\cite{luan2018multi,luan2019general}. The multiple tasks include concept recognition, relation extraction, and co-reference resolution.

Researchers notice label mistakes in many NLP tasks \cite{manning2011part,wang2019crossweigh, eskin2000detecting, kveton-oliva-2002-semi}. For instance, it is reported that the bottleneck of the POS tagging task is the consistency of the annotation result \cite{manning2011part}. People tried to detect label mistakes automatically and minimize the influence of noise in training. The mistake re-weighting mechanism is effective in the NER task \cite{wang2019crossweigh}. We focus on visually evaluating the label consistency.

\section{Conclusion}
\label{sec:conclusion}

We presented an empirical method to explore the relationship between label consistency and NER model performance. It identified the label inconsistency of test data in SCIERC and CoNLL03 datasets (with 26.7\% and 5.4\% label mistakes). It validated the label consistency in multiple sets of NER data annotation on two benchmarks, CoNLL03 and SCIERC.

\section*{Acknowledgements}
This work is supported by National Science Foundation CCF-1901059.

\balance
\bibliography{paper}

\begin{thebibliography}{14}
\expandafter\ifx\csname natexlab\endcsname\relax\def\natexlab#1{#1}\fi

\bibitem[{Eskin(2000)}]{eskin2000detecting}
Eleazar Eskin. 2000.
\newblock Detecting errors within a corpus using anomaly detection.
\newblock In \emph{Proceedings of the 1st North American chapter of the
  Association for Computational Linguistics conference}, pages 148--153.
  Association for Computational Linguistics.

\bibitem[{Jiang et~al.(2020)Jiang, Zeng, Zhao, Qin, Liu, Chawla, and
  Jiang}]{jiang2020biomedical}
Tianwen Jiang, Qingkai Zeng, Tong Zhao, Bing Qin, Ting Liu, Nitesh~V Chawla,
  and Meng Jiang. 2020.
\newblock Biomedical knowledge graphs construction from conditional statements.
\newblock \emph{IEEE/ACM transactions on computational biology and
  bioinformatics}, 18(3):823--835.

\bibitem[{Kv{\u{e}}to{\v{n}} and Oliva(2002)}]{kveton-oliva-2002-semi}
Pavel Kv{\u{e}}to{\v{n}} and Karel Oliva. 2002.
\newblock \href {https://www.aclweb.org/anthology/C02-1021} {(semi-)automatic
  detection of errors in {P}o{S}-tagged corpora}.
\newblock In \emph{{COLING} 2002: The 19th International Conference on
  Computational Linguistics}.

\bibitem[{Lample et~al.(2016)Lample, Ballesteros, Subramanian, Kawakami, and
  Dyer}]{lample2016neural}
Guillaume Lample, Miguel Ballesteros, Sandeep Subramanian, Kazuya Kawakami, and
  Chris Dyer. 2016.
\newblock Neural architectures for named entity recognition.
\newblock \emph{arXiv preprint arXiv:1603.01360}.

\bibitem[{Liu et~al.(2018)Liu, Shang, Ren, Xu, Gui, Peng, and
  Han}]{liu2018empower}
Liyuan Liu, Jingbo Shang, Xiang Ren, Frank~Fangzheng Xu, Huan Gui, Jian Peng,
  and Jiawei Han. 2018.
\newblock Empower sequence labeling with task-aware neural language model.
\newblock In \emph{Thirty-Second AAAI Conference on Artificial Intelligence}.

\bibitem[{Luan et~al.(2018)Luan, He, Ostendorf, and Hajishirzi}]{luan2018multi}
Yi~Luan, Luheng He, Mari Ostendorf, and Hannaneh Hajishirzi. 2018.
\newblock Multi-task identification of entities, relations, and coreference for
  scientific knowledge graph construction.
\newblock \emph{Proceedings of the 2018 Conference on Empirical Methods in
  Natural Language Processing (EMNLP)}.

\bibitem[{Luan et~al.(2019)Luan, Wadden, He, Shah, Ostendorf, and
  Hajishirzi}]{luan2019general}
Yi~Luan, Dave Wadden, Luheng He, Amy Shah, Mari Ostendorf, and Hannaneh
  Hajishirzi. 2019.
\newblock A general framework for information extraction using dynamic span
  graphs.
\newblock \emph{Proceedings of the 2019 Conference of the North American
  Chapter of the Association for Computational Linguistics: Human Language
  Technologies}.

\bibitem[{Manning(2011)}]{manning2011part}
Christopher~D Manning. 2011.
\newblock Part-of-speech tagging from 97\% to 100\%: is it time for some
  linguistics?
\newblock In \emph{International conference on intelligent text processing and
  computational linguistics}.

\bibitem[{Sang and De~Meulder(2003)}]{sang2003introduction}
Erik~F Sang and Fien De~Meulder. 2003.
\newblock Introduction to the conll-2003 shared task: Language-independent
  named entity recognition.
\newblock \emph{arXiv preprint cs/0306050}.

\bibitem[{Wang et~al.(2019)Wang, Shang, Liu, Lu, Liu, and
  Han}]{wang2019crossweigh}
Zihan Wang, Jingbo Shang, Liyuan Liu, Lihao Lu, Jiacheng Liu, and Jiawei Han.
  2019.
\newblock Crossweigh: Training named entity tagger from imperfect annotations.
\newblock In \emph{Proceedings of the 2019 Conference on Empirical Methods in
  Natural Language Processing and the 9th International Joint Conference on
  Natural Language Processing (EMNLP-IJCNLP)}.

\bibitem[{Yu et~al.(2020)Yu, Yu, Zhao, and Jiang}]{yu2020identifying}
Wenhao Yu, Mengxia Yu, Tong Zhao, and Meng Jiang. 2020.
\newblock Identifying referential intention with heterogeneous contexts.
\newblock In \emph{Proceedings of The Web Conference 2020}.

\bibitem[{Zeng et~al.(2021)Zeng, Lin, Yu, Cleland-Huang, and
  Jiang}]{zeng2021enhancing}
Qingkai Zeng, Jinfeng Lin, Wenhao Yu, Jane Cleland-Huang, and Meng Jiang. 2021.
\newblock Enhancing taxonomy completion with concept generation via fusing
  relational representations.
\newblock In \emph{ACM SIGKDD International Conference on Knowledge Discovery
  \& Data Mining (KDD)}.

\bibitem[{Zeng et~al.(2019)Zeng, Yu, Yu, Xiong, Shi, and
  Jiang}]{zeng2019faceted}
Qingkai Zeng, Mengxia Yu, Wenhao Yu, Jinjun Xiong, Yiyu Shi, and Meng Jiang.
  2019.
\newblock Faceted hierarchy: A new graph type to organize scientific concepts
  and a construction method.
\newblock In \emph{Proceedings of the Thirteenth Workshop on Graph-Based
  Methods for Natural Language Processing (TextGraphs-13)}.

\bibitem[{Zeng et~al.(2020)Zeng, Yu, Yu, Jiang, Weninger, and
  Jiang}]{zeng2020tri}
Qingkai Zeng, Wenhao Yu, Mengxia Yu, Tianwen Jiang, Tim Weninger, and Meng
  Jiang. 2020.
\newblock Tri-train: Automatic pre-fine tuning between pre-training and
  fine-tuning for sciner.
\newblock In \emph{Proceedings of the 2020 Conference on Empirical Methods in
  Natural Language Processing: Findings}.

\end{thebibliography}
\bibliographystyle{acl_natbib}












\end{document}